# Adversarial Example Defenses: Ensembles of Weak Defenses are not Strong

Warren He　　James Wei　　Xinyun Chen　　Nicholas Carlini　　Dawn Song
*UC Berkeley*　　*UC Berkeley*　　*UC Berkeley*　　*UC Berkeley*　　*UC Berkeley*

## Abstract

Ongoing research has proposed several methods to defend neural networks against adversarial examples, many of which researchers have shown to be ineffective. We ask whether a strong defense can be created by combining multiple (possibly weak) defenses. To answer this question, we study three defenses that follow this approach. Two of these are recently proposed defenses that intentionally combine components designed to work well together. A third defense combines three independent defenses. For all the components of these defenses and the combined defenses themselves, we show that an adaptive adversary can create adversarial examples successfully with low distortion. Thus, our work implies that ensemble of weak defenses is not sufficient to provide strong defense against adversarial examples.

## 1 Introduction

Neural networks have achieved great performance on a wide range of application domains; in particular, they have demonstrated accuracies comparable or better than humans on datasets in the fields of image recognition [10] and speech recognition [29]. However, recent work shows that deep learning models are susceptible to *adversarial examples*: inputs that are similar to a correctly classified input, but which are misclassified [28]. Research on other applications of neural networks has also encountered adversarial examples on different tasks beyond image classification, including deep policies in reinforcement learning and generative models [15, 14, 12, 2]. Adversarial examples pose serious threat in particular in security-critical autonomous systems such as self-driving cars. Recent work has shown that adversarial examples remain even when subject to the lossy channel of being photographed [17].

Developing effective defenses against adversarial examples is an important topic. Despite many attempts [27, 7, 9, 21, 8, 6, 5, 11], there is no strong defense against adversarial examples to date.

In this paper, we ask the question: *if we ensemble multiple defenses to adversarial examples, will the combined defense be significantly stronger than the original individual defense?* If it did, then one possible approach to constructing a robust defense to adversarial examples would be to join together many defenses, each of which independently is only slightly effective, but together are strong. This is an important question for designing effective defense against adversarial examples. To the best of our knowledge, we are the first to systematically investigate this question.

Towards answering this question, we study three instances of ensembled defenses. First, we study two recently proposed defenses, feature squeezing [30] and the *specialists+1* ensemble method [1], each of which ensembles multiple defenses that compensate for each other's weaknesses. Note that feature squeezing and the specialists+1 ensemble are explicitly designed to combine component defenses that work well together, with the intention of creating a stronger defense.

To study the question of ensemble defense in a broader scope, we also evaluate an ensemble of three independent, mutually compatible detection mechanisms [6, 21, 5]. This represents an approach that combines defenses which were not designed to be used together.

The works that introduce these defenses showed that they are effective at detecting attacks generated for the classifier models to which they are applied. However, we find that neither the components of these defenses nor the combined defenses are effective against an attacker that is aware of the defense.

We evaluate these defenses with new attacks, specific to the defenses. Our attacks are able to defeat all aforementioned defenses with low distortion. From this, we conclude that combining weak defenses, even ones seem to work well together, is insufficient for defending against an adaptive attacker.

**Contributions** We make the following contributions:

- We create effective attacks on feature squeezing [30], including individual squeezing methods and the combined adversarial example detection scheme.

- We create an effective attack on the ensemble-of-specialists defense [1].

- We create effective attacks on an ensemble of recently proposed detectors. We show that adversarial examples can bypass an ensemble of detectors with nearly as little distortion as needed for the strongest individual detector.

- Our results show that ensembled defenses do not provide significantly more resilience against adversarial examples than each individual component included in the ensemble. This implies that an ensemble of weak defenses is not sufficient to provide a strong defense against adversarial examples.

- Our evaluation demonstrates that adaptive adversarial examples transfer across several defense or detection proposals. This phenomenon may provide one reason to explain why ensembling is not an effective approach to building defense mechanisms against adversarial examples.

The rest of this paper is organized as follows: in Section 2, we provide the problem statement and background information; in Section 3, we describe our attacks on individual feature squeezing defense component and their composite defense; in Section 4, we describe our attack results on the ensemble-of-specialists defense; in Section 5, we describe our attack results on a composite defense that combines multiple independently proposed detection networks; and we summarize our findings in Section 6.

## 2 Overview

We start with an overview of background information, then we define the threat models we use and the problem statement and setup for our experiments.

### 2.1 Background: Adversarial Examples

Recent works have pointed out that deep learning models are vulnerable to adversarial examples: these models give incorrect predictions on inputs that are slightly different from those correctly predicted ones [28, 7, 23, 26].

Specifically, suppose we have a classifier $F$ with model parameters $\theta$. Let $x$ be an input to the classifier with corresponding ground truth prediction $y$. An adversarial example $x^*$ is some instance in the input space that is close to $x$ by some distance metric $d(x, x^*)$, but causes $F_\theta$ to produce an incorrect output. Here we only consider those $x$ satisfying $F_\theta(x) = y$.

Prior work considers two classes of adversarial examples. First, an *untargeted* adversarial example is an instance $x^*$ that causes the classifier to produce any incorrect output: $F_\theta(x^*) \neq y$. Second, a *targeted* adversarial example is an $x^*$ that causes the classifier to produce a specific incorrect output $y^*$: $F_\theta(x^*) = y^*$ where $y \neq y^*$.

Several approaches have been proposed in previous work, including the Fast Gradient Sign Method (FGSM) [7], Fast Gradient Method [19], Jacobian-based Saliency Map Approach (JSMA) [26], Deepfool [22], and optimization-based methods [28, 3, 19].

### 2.2 Threat Models

In this work, we assume the adversary has full knowledge of the model, including the model architecture, parameters, and the defense strategies used in the model. Prior work has shown this assumption can often be relaxed [7, 24, 25], however for simplicity we assume this stronger threat model.

Within these white-box attackers, we consider two capacities of adversaries.

**Static Adversary.** A *static adversary* is an attacker that is not aware of any defenses that may be in place to protect the model against adversarial examples. A static adversary can generate adversarial examples using existing methods but does not tailor attacks to any specific defense.

**Adaptive Adversary.** An *adaptive adversary* is an attacker that is aware of the defense methods used in the model, and can adapt attacks accordingly. This is a strictly more powerful adversary than static adversary. In this paper, we consider this stronger adversary.

### 2.3 Problem Statement

To improve the robustness of models against adversarial examples, prior work investigates into two directions. The first direction attempts to produce correct predictions on adversarial examples, while not compromising the accuracy on legitimate inputs [27, 7, 9]. The other (more recent) direction instead attempts to *detect* adversarial examples, without introducing too many false positives. In this case, the model can reject an instance and refuse to classify those that it detects as adversarial [21, 8, 30, 1].

In this paper, we ask the question: *if we ensemble multiple defenses to adversarial examples, then will the combined defense be significantly stronger than each individual original defense?* If it did, then one possible approach to constructing a robust defense to adversarial examples would be to join together many defenses, each of which independently is only slightly effective, but together are strong. This is an important question for designing effective defense against adversarial examples. To the best of our knowledge, we are the first to systematically investigate this question.

**Defenses considered.** In this paper, we consider defenses that attempt to combine multiple (somewhat weaker) defenses to construct a larger strong defense. In particular, we look at three instances of ensemble defense strategies. First and second are feature squeezing [30] and the *specialists+1* ensemble method [1], both of which take this approach by construction. These defenses are constructed from components that are intended to be useful together. Their authors have shown that these defenses effectively detect low-perturbation adversarial examples generated by a static adversary. Third, to study the effectiveness of ensembling defenses more broadly, we merge together many detectors that were not designed to be used in conjunction with any other detector. In particular, as an example demonstration, we ensemble three independent detection mechanisms [6, 21, 5] to build one detection mechanism.

For each of these defense strategies, we propose attack methods to generate adversarial examples as an adaptive adversary against the individual component defense (when applicable) as well as the composite defense strategy. We use these attack methods to evaluate each component defense and composite defense: if our method succeeds at generating adversarial examples, this means that an adaptive adversary can defeat the defense. To gauge how strong the combined defense is compared to the components, we compare the level of distortion needed to fool each (using the same optimization method).

## 2.4 Experimental Setup

**Datasets and models.** To evaluate the effectiveness of the different defense strategies, we use two standard datasets, MNIST [18] and CIFAR-10 [16] datasets.

For both datasets, we randomly sample 100 images in the test set, filter out examples that are not correctly classified, and generate adversarial examples based on the correctly classified images. When evaluating each defense strategy, we use the same model architectures described in their papers respectively [30, 1, 6, 21, 5].

Adversarial examples on MNIST tend to have higher distortion than natural images.

**Adversarial example generation method.** In this paper, we use an optimization-based approach to generate adversarial examples [3], which is shown to be effective on finding adversarial examples with small distortions.

At a high level, the attack uses an optimizer to minimize a loss function:

$$\text{loss}(x') = \|x' - x\|_2^2 + c \cdot J(F_\theta(x'), y)$$

Here, $F_\theta$ is a part of the trained classifier that outputs a vector of logits, and $J$ computes some penalty based on the logits and some label $y$, either a ground truth label for non-targeted attacks or a target label for targeted attacks. A constant $c$ is a hyperparameter that adjusts the relative weighting between distortion and misclassification. We omit details of the design choice and refer the reader to the original paper [3].

**Measurement of distortion.** Unless otherwise specified, we measure the distortion between an adversarial example and the original input as the $L_2$-norm of their distance. Formally, $d(x^*, x) = \sqrt{\sum_i (x_i^* - x_i)^2}$. Each dimension of input images is scaled to $[0, 1]$, i.e., $0 \leq x_i^*, x_i \leq 1$.

## 3 Adaptive attacks on feature squeezing

In this and next section, we investigate ensemble defense strategies that are intentionally constructed to have component defenses which work together to detect adversarial examples. The first defense we study is *feature squeezing*, proposed by Xu et al. [30, 31].

**Background: feature squeezing defense.** To perform feature squeezing, one generates a lower fidelity version of the input image through a process known as "squeezing" before passing it into the classifier. Xu et al. proposed two methods of squeezing: reducing the color depth to fewer bits, and spatially smoothing the pixels with a median filter. According to their paper, the two methods of squeezing work well together because they address two major kinds of perturbation used in adversarial examples: color depth reduction eliminates small changes to many pixels, while spatial smoothing eliminates large changes to a few pixels.

In order to detect adversarial examples, Xu et al. propose a system combining the two squeezing methods. First, the system runs the classifier on three different versions of the image: the original image, the reduced-color-depth version and the spatially smoothed version of the original image. Then, it compares the softmax probability vectors across these three classifier outputs. The $L_1$ *score* of the input is the highest $L_1$ distance between any pair of softmax probability vectors among the three. It flags inputs where the $L_1$ score exceeds a threshold as adversarial.

In their experiments, Xu et al. show that MNIST and CIFAR-10 classifiers are accurate on squeezed inputs. On adversarial examples generated by a static adversary using FGSM [7] and JSMA [26], they show that their detector achieves 99.74% accuracy on a test set with equal portions benign and adversarial examples. They also show that squeezing the input alone prevents 84 – 100% of the adversarial examples (correctly classifying them). Recently, Xu et al. showed that a simplified detector that uses the original version of the input and the spatially-smoothed version (excluding the color-depth-reduced version) achieves a 98.80% overall detection accuracy on MNIST and 87.50% on CIFAR-10 against a static adversary using a variety of Carlini and Wagner's attacks [31].

**Summary of our approach and results.** We demonstrate that feature squeezing is not an effective defense in two stages. First, we show that an adaptive attacker can construct an adversarial example that remains adversarial after it is squeezed by each method (color depth reduction and spatial smoothing. Then, we use this approach to construct adversarial examples that are classified the same way both with and without squeezing, causing the $L_1$ score to be smaller than a given fixed threshold. Our results show that the combined detection method is not effective against an adaptive attacker.

## 3.1 Evading individual feature squeezing defense components

In these experiments, we evaluate whether adversarial examples are robust to each individual feature squeezing defense component, i.e., whether adversarial examples remain adversarial after each individual feature squeezing process (color depth reduction and spatial smoothing) separately. These experiments attack the components of the combined feature squeezing detection scheme. Performing this attack is necessary for defeating the combined detection scheme, wherein the predicted label probabilities of squeezed images are compared against each other.

### 3.1.1 Evading color-depth-reduction defense

The first method of squeezing an image that Xu et al. propose is color depth reduction. This method rounds each value in the input to $2^b$ evenly spaced values spanning the same range, which we refer to as reducing to *b* bits.

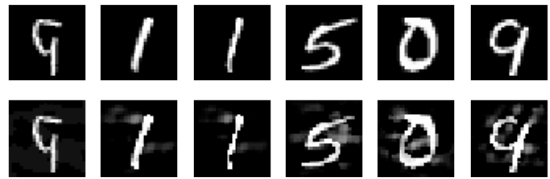

Figure 1: Adversarial examples for color depth reduction (to 1 bit) on MNIST. First row: original images. Second row: adversarially perturbed. Distortions, from left to right: 1.49, 2.61, 2.63, 3.83, 3.89, 3.90.

**Attack Approach.** We use the method described in Section 2.4 to generate adversarial examples that are robust to color depth reduction. After each step of the optimization procedure, an intermediate image (perturbed from the original image) is available from the optimizer. We check if a reduced-color-depth version of this intermediate image is adversarial. We run the optimization multiple times, initializing the optimization with random perturbations of the original image each time, so that it explores different optimization paths. For each original image, we keep the successful adversarial example that has the lowest $L_2$ distance to the original image among all the generated successful adversarial examples for this original image.

Although this approach successfully generated low-distortion adversarial examples in our experiments, there is no guarantee that it should succeed in the general case. We present an alternative approach in Appendix A which produced examples with higher distortion, but which may be useful in other scenarios.

**Attack results on MNIST.** We evaluate color depth reduction to 1 – 7 bits. On the strongest defense evaluated by Xu et al., which reduces color depth to 1 bit, we successfully generated adversarial examples for *all* original images, with an average distortion of 3.86. Figure 1 shows a sample of these adversarial examples. Table 1 summarizes our results for other bit depths. Notice that for a system without any color depth reduction (retaining the original 8 bits of depth), we find adversarial examples with an average distortion of 1.38. Reducing color depth to fewer bits makes the system less sensitive to small changes, which requires larger distortions; however, the distortions are still very small.

**Attack results on CIFAR-10.** We evaluate color depth reduction to 3 bits, which Xu et al. recommend as a good balance between the accuracy on adversarial inputs and accuracy on benign images for CIFAR-10. We succeeded at generating adversarial examples for *all* original images, with an average distortion of 0.945. Figure 2 shows a sample of these adversarial examples. For comparison, adversarial examples for a classifier without color depth reduction have an average distortion of 0.214. Although this method of squeezing increases the distortion needed for successfully generating non-targeted adversarial examples using the same optimization method, again, such a distortion is still small and imperceptible.

| Bit depth | Adv success | Avg $L_2$ |
|---|---|---|
| 1 | 100% | 3.86 |
| 2 | 99% | 1.69 |
| 3 | 100% | 1.43 |
| 4 | 100% | 1.39 |
| 5 | 100% | 1.44 |
| 6 | 100% | 1.33 |
| 7 | 100% | 1.33 |
| 8 | 100% | 1.38 |

Table 1: Summary of MNIST adversarial examples that are misclassified when reduced to different color depths. "Adv success" measures the fraction of original images for which we successfully found an adversarial example. "Avg $L_2$" measures the average $L_2$ distortion of the successful adversarial examples.

| Filter size | Adv success | Avg $L_2$ |
|---|---|---|
| $3 \times 3$ | 100% | 1.29 |
| $2 \times 2$ | 100% | 1.57 |
| $5 \times 5$ | 100% | 0.612 |
| $3 \times 1$ | 100% | 1.33 |
| $1 \times 3$ | 100% | 1.29 |
| $2 \times 1$ | 100% | 1.52 |
| $1 \times 2$ | 100% | 1.51 |
| $5 \times 1$ | 100% | 0.943 |
| $1 \times 5$ | 100% | 0.931 |

Table 2: Summary of MNIST adversarial examples that are misclassified when spatially smoothed with varying sizes of median filters. Columns have the same meaning as in Table 1. Some filters make adversarial examples *easier* to find.

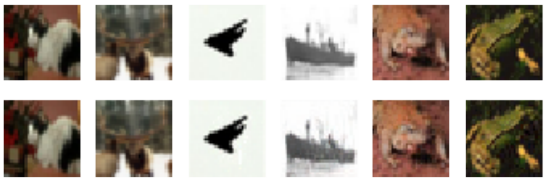

Figure 2: Adversarial examples for color depth reduction (to 3 bits) on CIFAR-10. Distortions, from left to right: 0.0194, 0.0954, 0.322, 0.942, 0.948, 0.948. Layout is the same as Figure 1.

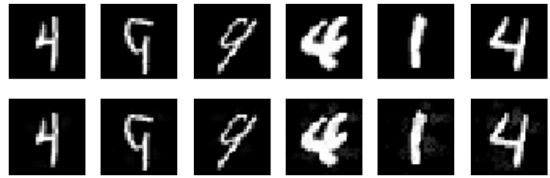

Figure 3: Adversarial examples for spatial smoothing (with $3 \times 3$ filter) on MNIST. Distortions, from left to right: 0.236, 0.241, 0.282, 1.27, 1,31, 1.31. Layout is the same as Figure 1.

**Summary.** An adaptive attacker can successfully generate adversarial examples with small distortions for a system that applies color depth reduction to the input image before classifying it.

### 3.1.2 Evading spatial smoothing

Xu et al. propose a second method for feature squeezing, which applies a median filter to the input, which replaces each pixel with the median value of a neighborhood around the pixel.

To generate adversarial examples that are misclassified after spatial smoothing, we use the procedure from Section 2.4 with the addition of a median filter as part of the classification model. A median filter for TensorFlow was not available, so we implemented our own.

**Attack results on MNIST.** We evaluate a range of median filter sizes, ranging from $1 \times 2$ to $5 \times 5$. For a $3 \times 3$ filter, with which Xu et al. achieved the best accuracy, we successfully generated adversarial examples for *all* original images, with an average distortion of 1.29. Figure 3 shows a sample of these adversarial examples. Table 2 summarizes our results for other filter sizes. Larger median filters did not require greater distortion. Compared to adversarial examples generated for a system without any spatial smoothing (average distortion of 1.38), the average distortion is not increased.

**Attack results on CIFAR-10** We evaluate a $2 \times 2$ median filter, which Xu et al. identify as achieving a good rejection rate of adversarial examples and accuracy on benign images on CIFAR-10. We successfully generated adversarial examples for *all* original images, which have an average distortion of 0.205. Figure 4 shows a sample of these adversarial examples. The average distortion is not higher than for a system without spatial smoothing (0.214).

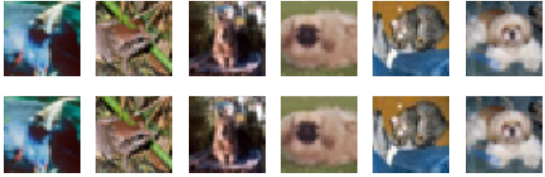

Figure 4: Adversarial examples for spatial smoothing (with 2 × 2 filter) on CIFAR-10. Distortions, from left to right: 0.0273, 0.0537, 0.0584, 0.198, 0.211, 0.212. Layout is the same as Figure 1.

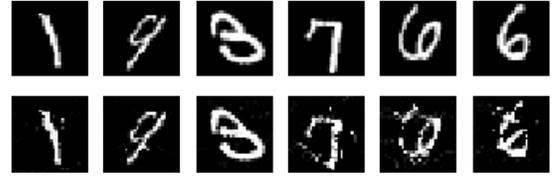

Figure 5: Adversarial examples for combined feature squeezing detection on MNIST. Distortions, from left to right: 2.00, 2.04, 2.39, 4.66, 4.77, 4.79. Layout is the same as Figure 1.

**Summary** Spatial smoothing alone is not an effective defense against an adaptive attacker. We have shown that an adaptive adversary can create adversarial examples for a system that applies spatial smoothing which are not more distorted than adversarial examples for a baseline system that does not apply spatial smoothing.

## 3.2 Evading combination of multiple squeezing techniques

While the individual feature squeezing techniques are weak against an adaptive attacker, we investigate whether the detection scheme that combines them is stronger. In this case, we find that this detection scheme is not much stronger than the strongest component defense, color depth reduction.

**Background: Composite feature squeezing defense** The detection scheme combines both methods of squeezing. In particular, the detection system has three "branches," where each one runs the classifier on a different version of the input, the original input, a reduced-color-depth version and a spatially-smoothed version of the original input. These three branches output different softmax probability vectors, and the scheme compares the $L_1$ distance between these vectors against a threshold to determine whether the input is adversarial.

In this experiment, we evaluate the effectiveness of the combined defense against an adaptive attacker, i.e., whether adversarial examples can be misclassified and bypass this detection scheme.

**Attack Approach.** We use the procedure from Section 3.1.1 with modifications to support the detection mechanism. In this attack, we run two instances of the model in parallel during optimization, representing the detector branches that operate on the input and the spatially smoothed version of the input. The optimizer receives gradients from both models, equally. We do not include the reduced-color-depth branch in the gradient calculation, because the color depth reduction removes gradient information; it is, however, included when we compute the $L_1$ score. We collect only adversarial examples that have an $L_1$ score below a threshold of 0.3076, a level at which Xu et al. achieved the best accuracy in their experiments on MNIST.

**Attack results on MNIST** We evaluate a combination of color depth reduction to 1 bit and smoothing with a 2 × 2 median filter, which Xu et al. found to be accurate on adversarial examples generated by a static adversary [31]. We successfully generated adversarial examples for *all* original images, with an average distortion of 4.76 and $L_1$ score of 0.209. Figure 5 shows a sample of these adversarial examples. These examples are misclassified and successfully evade detection. This distortion is 23.3% larger than for color depth reduction alone, but still very small.

**Attack results on CIFAR-10.** We evaluate a combination of color depth reduction to 3 bits and smoothing with a 2 × 2 median filter, a combination of settings that perform well in Xu et al.'s experiments. We successfully generated adversarial examples for *all* original images, with an average distortion of 0.601 and $L_1$ score of 0.168. Figure 6 shows a sample of these adversarial examples. These examples are misclassified and successfully evade detection. This distortion is even lower than that of the color depth reduction defense alone. Although Xu et al. do not prescribe a threshold specific to CIFAR-10, the average $L_1$ score for these examples is lower (i.e., detected as less adversarial) than the average $L_1$ score for the original images, which is 0.225.

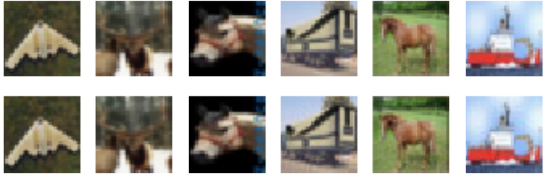

Figure 6: Adversarial examples for combined feature squeezing detection on CIFAR-10. Distortions, from left to right: 0.117, 0.120, 0.130, 0.604, 0.614, 0.617. Layout is the same as Figure 1.

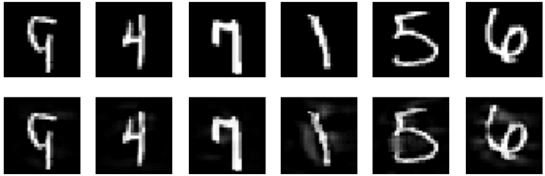

Figure 7: Adversarial examples for specialists+1 on MNIST. Distortions, from left to right: 0.587, 0.659, 0.808, 2.49, 2.51, 2.51. Layout is the same as Figure 1. Results show that distortion in generated adversarial examples are imperceptible.

**Summary.** The detection scheme that combines two methods of squeezing is not always stronger than the strongest component, color depth reduction. The improvement is low even on MNIST, which is particularly well suited for feature squeezing, with images being black and white (little change from color depth reduction) and having large, contiguous areas of the same color (little change from spatial smoothing). On CIFAR-10, the combined attack requires less distortion than the color depth reduction defense alone.

## 4 Evading ensemble of specialists

We study a second defense that combines multiple component defenses, an ensemble of specialists, proposed by Abbasi and Gagné [1].

**Background: ensemble of specialist defense.** The defense consists of a generalist classifier (which classifies among all classes) and a collection of specialists (which classify among subsets of the classes). The specialists classify subsets of the classes as follows. Where $C$ is the set of all $K$ classes in the task, for each class $i$, let $U_i$ be the set of classes with which $i$ is most often confused in adversarial examples. To compute $U_i$, Abbasi and Gagné select the top 80% of misclassifications caused by non-targeted FGSM attacks for each class $i$. Further, $K$ additional subsets are defined: $U_{K+i} = C \setminus U_i$ to be the complement set of $U_i$. For each $j = 1, ..., 2K$, a specialist classifier $F_j$ is trained on a subset of the dataset containing images belonging to the classes in $U_j$ to classify input images into the classes in $U_j$ only. In addition, a generalist classifier $F_{2K+1}$ is trained to classify input images into classes in $U_{2K+1} = C$. Each classifier in the ensemble may be susceptible to basic adversarial examples, but the proposed defense assumes that each specialist can detect a few specific attacks, thus the attacker cannot fool all specialists and the generalist at the same time. The defense combines them to jointly detect general adversarial examples.

In order to classify an input, the system first checks if, for any class $i$, the generalist classifier and all specialists that can classify $i$ agree that the input belongs to class $i$. If such a class $i$ exists, note that at most one class can get the generalist's vote, it must be unique. In this case, the system outputs $i$ as the class. Otherwise, at least one classifier has misclassified the input, and the prediction follows the majority vote from all classifiers in the ensemble.

Abbasi and Gagné [1] find that using an ensemble constructed this way successfully reduces the system's confidence (fraction of voting classifiers that voted for the winning class) on adversarial examples generated by a static attacker using FGSM [7], DeepFool [22], and Szegedy et al.'s approach [28]. They conclude that a classification system can use an ensemble of diverse specialists this way and detect low-confidence examples as adversarial.

**Attack approach.** In this experiment, we evaluate the effectiveness of Abbasi and Gagné's specialists+1 ensemble against an adaptive attacker. We considered a scenario where a user provides an image to a system, and the system uses a specialists+1 ensemble to classify the image or reject it as adversarial.

We attempt to create targeted adversarial examples, where we chose target classes randomly. For each original image, then our goal is to create an adversarial example that is classified as the target class by the generalist classifier and all applicable specialists at the same time, resulting in the maximum possible confidence that the image is not adversarial. We use the procedure from Section 2.4 to generate adversarial examples. We modified the loss function to support multiple classifiers:

$$\text{loss}(x') = \|x' - x\|_2^2 + c \sum_{j \in \{1,...,2K+1\}; y^* \in U_j} J(F_j(x'), y^*)$$

We evaluate this defense on MNIST only. While Abbasi and Gagné also propose the defense for CIFAR-10,

the architecture described in their experiments have low accuracy, resulting in low confidence even in benign images.

**Attack results on MNIST.** We successfully generated adversarial examples for *all* original images, which have an average distortion of 2.50. Figure 7 shows a sample of these adversarial examples in the second row. These adversarial examples are classified as the target label by the generalist and all applicable specialists, giving them the highest confidence possible. The distortion needed is 81.2% higher than for a non-ensemble MNIST classifier.

Although this defense defines the specialists to focus on common misclassifications caused by non-targeted adversarial examples, it is still weaker at detecting the common misclassifications. Among the examples, 33 targeted a class that the original image's ground truth class was commonly confused with. The average distortion for these images is 1.86, below the average of the entire set.

**Summary.** The specialists+1 ensemble does not effectively ensure low confidence on adversarial examples generated by an adaptive attacker. An adaptive attacker can successfully generate adversarial examples with small distortions, which are unanimously classified as a target class, and thus evade the detection of the specialist+1 ensemble defense.

## 5 Evading ensemble of detectors

In the previous sections, we have investigated ensembles of defenses that are intentionally constructed to be useful together. In Xu et al.'s work, the color depth reduction is intended to remove small changes to many pixels, and the median smoothing to remove large changes to a few pixels. Similarly, Abbasi and Gagné propose using an ensemble of generalist and specialist classifiers together; without the others, this approach would not be useful.

To study the effectiveness of ensembling defenses more broadly, we merge together a number of recently proposed detectors that were not designed to be used in conjunction with any other detector. We consider only detectors that are applied to a fixed classification network for simplicity, and therefore study the following schemes:

- Gong et al. propose using adversarial training to detect adversarial examples [6]. Given the original model, generate adversarial examples on the training data. Then, train a new classifier that distinguishes the original training data from the adversarial data.

|        | Source Defense |        |         |
|--------|------|--------|---------|
|        | Gong | Metzen | Feinman |
| Gong (Target)   | 100% | 51%    | 21%     |
| Metzen (Target) | 43%  | 100%   | 18%     |
| Feinman (Target)| 96%  | 92%    | 100%    |

Table 3: Probability that adversarial examples constructed for a given source defense also fool the given target defense on CIFAR-10. Defenses generated against Metzen *et al.* transfer to the others with the highest probability, and Feinman *et al.* with lowest.

- Metzen et al. construct a similar scheme, however instead of using the original images as the input to the detector, they train on the inner convolutional layers of the network [21].

- Feinman et al. examine the final hidden layer of a neural network and find that adversarial examples are separable from the original images by training a density estimate using Gaussian kernels [5].

When using Carlini and Wagner's attack, these approaches are known to provide only slight increases in robustness, i.e., only increase the required distortion slightly when generating the adversarial examples with the detector vs. without the detector [4]. Given this, we now examine if constructing an ensemble of these defenses provides additional robustness. To ensemble these defenses, we run each detection method and report the input as adversarial if any of the three detectors do.

**Attack approach.** We perform this experiment on CIFAR-10 exclusively, as Metzen et al.'s defense is intended for CIFAR-10. We are able to construct adversarial examples for all defenses independently. To defeat all three defenses together, we construct a new classifier $G(\cdot)$ so that using the loss function from Section 2.4 directly can construct adversarial examples.

We use the same notation as [4]. Let $F(\cdot)$ be a classifier on $N$ classes, and softmax$(F(\cdot))_i$ be the probability of class $i$ (so that $F(\cdot)_i$ are the logits). Let $\{D_j(x)\}_{j=1}^{J}$ be one of $J$ different detectors so that the probability that detector $D_j$ reports object $x$ as adversarial is sigmoid$(D_j(x))$ (that is, $D_j$ returns the logits). We report that an instance is an adversarial example if the probability of any detector is greater than one half. That is, if for any $j$, sigmoid$(D_j(x)) > \frac{1}{2}$, or, alternatively, $D_j(x) > 0$.

When we ensemble the three defenses, we set $J = 3$ and define $D(x) = \max_j D_j(x)$, so that $D(x)$ reports adversarial (i.e., $D(x) > 0$) if any of the three detectors do.

Given this, we use the same $G(\cdot)$ construction as Carlini and Wagner's previous work on these de-fenses [4]. This function $G(\cdot)$ returns N+1 classes (with the new class reserved for adversarial examples) so that $\arg\max_i G(x)_i = \arg\max_i F(x)_i$ when $x$ is not adversarial, and $\arg\max_i G(x) = N+1$ when $x$ is adversarial. To do this, [4] specifically defines

$$G(x)_i = \begin{cases} F(x)_i & \text{if } i \leq N \\ (D(x)+1) \cdot \max_j F(x)_j & \text{if } i = N+1 \end{cases}$$

If for a given instance $x$, $D_j(x) > 0$ (for any classifier $j$) then we will have $\arg\max_i G(x)_i = N+1$ since we multiply a value greater than one by the largest of the other output logits. Conversely, if $\arg\max_i G(x)_i \neq N+1$ then we must have $D(x) < 0$ implying that all detectors report the instance is benign.

Therefore, by constructing adversarial examples on $G$ so that the target class is not $N+1$, we can construct adversarial examples on $F$ that are not detected by any detector.

**Attack results on CIFAR-10** The $L_2$ distortion required to construct adversarial examples on an unsecured network is 0.11. To construct adversarial examples on this network $G(\cdot)$ with the three defenses increases the distortion to 0.18, an increase of 60%. However, this distortion is still imperceptible.

**Transferability of adversarial examples across different detectors.** In order to understand the reason that these defenses do not significantly increase robustness when combined together, we hypothesize that the transferability property [28, 7, 24, 19] of adversarial examples is simplifying the attacker's task. To verify this, we construct adversarial examples on each of the three defenses in isolation and check the probability that these examples also fool the other two defenses. Table 3 contains this data. From this, we can see why constructing an ensemble of these weak defenses is not significantly more secure than each independently: the adversarial examples that fool one detector also often fool the other detectors.

## 6 Conclusion

Our goal in this paper is to examine whether multiple (possibly weak) defenses can be combined to create a strong defense. Towards this goal, we studied three such defenses that combined multiple defense components: two recently proposed defenses designed with rationale of why their components should work well together and one that combined unrelated recently proposed detectors.

We showed that an adaptive adversary can generate adversarial examples with low distortion that fool all of the defenses and components that we evaluate. The feature squeezing detection scheme, which combines two methods of squeezing an input image, is at best marginally stronger than color depth reduction alone. The specialists+1 ensemble, which combines several specialist classifiers, increases the required distortion slightly, but again, distortion is still small. We also showed that combining a collection of recently proposed detection mechanisms is also ineffective. In particular, our results show that adversarial examples transfer across the individual detectors.

Our work sheds light on a few important lessons when evaluating defenses against adversarial examples: 1) one should evaluate defenses using strong attacks. For example, FGSM can quickly generate adversarial examples, but may fail to generate successful attacks when other iterative optimization based methods can succeed; 2) one should evaluate defenses using adaptive adversaries. It is important to develop defenses that are secure against attackers who know the defense mechanisms being used.

Finally, our results indicate that combining weak defenses does not significantly improve the robustness of these systems. Developing effective defenses against adversarial examples is an important topic. We hope our work sheds light for future work in this area.

## 7 Acknowledgements

We thank Arjun Bhagoji, Chang Liu, and Richard Shin, who have provided valuable feedback. This work was supported in part by BDD (Berkeley Deep Drive); the CLTC (Center for Long-Term Cybersecurity); FORCES (Foundations Of Resilient CybEr-Physical Systems), which receives support from the National Science Foundation (NSF award numbers CNS-1238959, CNS-1238962, CNS-1239054, CNS-1239166); and the National Science Foundation under Grant No. TWC-1409915. Any opinions, findings, and conclusions or recommendations expressed in this material are those of the author(s) and do not necessarily reflect the views of the National Science Foundation. Cloud computing resources were provided through a Microsoft Azure for Research award.

## A  Gumbel-Softmax reparameterization

Defenses that mask a network's gradients by quantizing the input values pose a challenge to gradient-based optimization methods for generating adversarial examples, such as the procedure we describe in Section 2.4. A straightforward application of the approach would find zero gradients, because small changes to the input do not alter the output at all. In Section 3.1.1, we describe an approach where we run the optimizer on a substitute network without the color depth reduction step, which approximates the real network. We rely on the optimizer stumbling across images that happen to be misclassified after the color depth reduction would take place. This worked in our experiments.

In this section, we describe an alternative approach that allows us to use the same optimizer (Adam) to search over the space of quantized images, by using *Gumbel-Softmax reparameterization* [13, 20].

**Background: Gumbel-Softmax reparameterization.** In short, the Gumbel-Softmax reparameterization trick changes a discrete optimization problem into a problem of optimizing distributions. In the reparameterized problem, the input image is created by sampling discrete-valued samples from distributions (one per channel, per pixel). Each distribution is represented by a set of continuous-valued parameters. During backpropagation, gradients arriving at the integer-valued samples affect the parameters representing the distribution from which the samples were drawn, pushing the distribution toward an optimal value.

| Bit depth | Adv success | Avg $L_2$ | $L_2$ bound |
|---|---|---|---|
| 1 | 100% | 7.49 | 2.57 |
| 2 | 100% | 6.38 | 0.920 |
| 3 | 100% | 5.72 | 0.407 |
| 4 | 100% | 5.36 | 0.197 |
| 5 | 100% | 5.33 | 0.108 |
| 6 | 100% | 5.34 | 0.0606 |
| 7 | 100% | 5.63 | 0.0287 |

Table 4: Summary of MNIST adversarial examples generated by Gumbel-Softmax reparameteriazation. Columns have the same meaning as in Table 1. "$L_2$ bound" measures the lowest possible average $L_2$ distance from integer-valued images to the original images, which is when the original images' pixels are rounded to the nearest quantized color.

**Attack approach.** We set up a network to sample a discrete-valued image, acting as a color-depth-reduced input, connected to the input of a classifier. The network samples the discrete-valued image from Gumbel-Softmax distributions (one distribution for each channel of each pixel). We the approach from Section 2.4, but optimizing the distributions instead of optimizing the image directly. At each step of the optimization, we check if the sampled image is misclassified, and we retain the one closest to the original image.

**Attack results on MNIST.** We evaluate color depth reduction to 1 – 7 bits. For reduction to 1 bit (the strongest color depth reduction evaluated), we successfully generated adversarial examples for *all* original images, with an average $L_2$ distortion of 7.48. For comparison, simply rounding the original images to 1-bit color depth results in an average $L_2$ distortion of 2.56. Table 4 summarizes our results for other bit depths. Again, squeezing to fewer bits increases the average distortion.

**CIFAR-10** We evaluated color depth reduction to 3 bits (a good balance between accuracy on adversarial inputs and on benign images). We succeeded at generating adversarial examples for *all* original images, with an average $L_2$ distortion of 2.32. For comparison, simply rounding the original images to 3-bit color results in an average distortion of 2.27.

**Summary** The reparameterized optimization succeeded at generating adversarial images, but the procedure did not produce adversarial images with the lowest distance to original (unquantized) images. This is, in part, because it always results in an image that is already quantized, which is necessarily at least some distance away from a high-color-depth original image. However, this approach may be useful in attacks on other defenses that try to obfuscate gradients through quantization. We find that the distance to the unquantized images is not much larger than the minimal distance between the original image and its normally quantized version.